\documentclass[letterpaper, 10 pt, conference]{ieeeconf}  

\IEEEoverridecommandlockouts                              

\usepackage[draft]{pgf}
\usepackage{cite}
\usepackage{balance}
\usepackage{graphics} 
\usepackage{graphicx} 
\usepackage{epsfig} 
\usepackage{amsmath, bm} 
\usepackage{amssymb}  
\usepackage{upgreek}
\usepackage{bbm, bbold}
\usepackage{algorithm}
\usepackage{algpseudocode}

\usepackage{cuted}
\usepackage{booktabs}
\usepackage{multicol}
\usepackage{multirow}
\usepackage{stfloats}
\usepackage{lipsum}  
\usepackage{breqn}
\usepackage{comment} 
\usepackage{accents}

\PassOptionsToPackage{hyphens}{url}\usepackage{hyperref}

\title{\LARGE \bf
Optimization-free Ground Contact Force Constraint Satisfaction in Quadrupedal Locomotion
}

\author{Eric Sihite, Pravin Dangol, and Alireza Ramezani$^{1}$
\thanks{$^{1}${\it SiliconSynapse Laboratory}, Electrical and Computer Engineering Department, Northeastern University, Boston, MA, USA.}
\thanks{emails: \{e.sihite, dangol.p, a.ramezani\}@northeastern.edu}%
}

\begin{document}
\maketitle

\global\csname @topnum\endcsname 0
\global\csname @botnum\endcsname 0
\thispagestyle{empty}
\pagestyle{empty}

\begin{abstract}

We are seeking control design paradigms for legged systems that allow bypassing costly algorithms that depend on heavy on-board computers widely used in these systems and yet being able to match what they can do by using less expensive optimization-free frameworks. In this work, we present our preliminary results in modeling and control design of a quadrupedal robot called \textit{Husky Carbon}, which under development at Northeastern University (NU) in Boston. In our approach, we utilized a supervisory controller and an Explicit Reference Governor (ERG) to enforce ground reaction force constraints. These constraints are usually enforced using costly optimizations. However, in this work, the ERG manipulates the state references applied to the supervisory controller to enforce the ground contact constraints through an updated law based on Lyapunov stability arguments. As a result, the approach is much faster to compute than the widely used optimization-based methods. 

\end{abstract}


\section{Introduction}
\label{sec:introduction}
 
\begin{figure}[!t]
    \vspace{0.1in}
    \centering
    \includegraphics[width = 1 \linewidth]{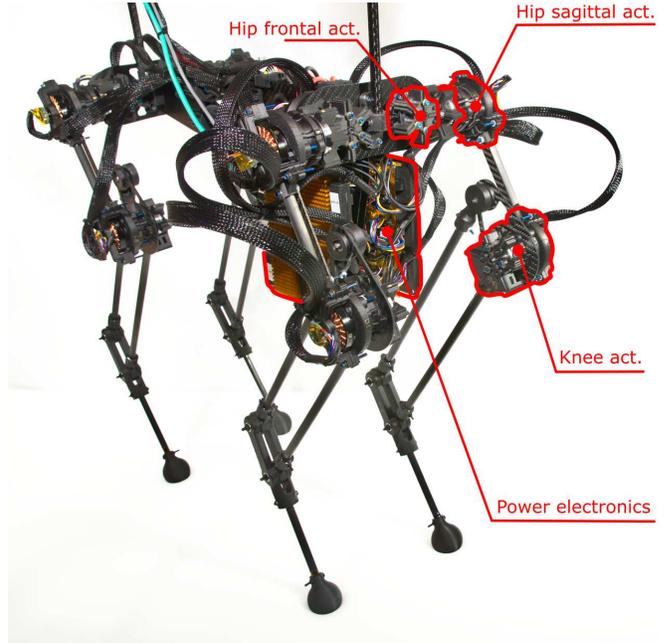}
    \caption{NU's Husky Carbon, the platform to integrate legged and aerial mobility.}
    \vspace{-0.1in}
    \label{fig:husky_overview}
\end{figure}

In this work, we will report our preliminary results in the modeling, control design, and development of a legged robot called \textit{NU's Husky Carbon} (shown in Fig.~\ref{fig:husky_overview}). Our objective is to integrate legged and aerial mobility into a single platform.

Robotic biomimicry of animals' multi-modal locomotion can be a significant ordeal. The prohibitive design restrictions such as tight power budget, limited payload, complex multi-modal actuation and perception, an excessive number of active and passive joints involved in each mode, sophisticated control, and environment-specific models, just to name a few, have alienated these concepts. It is worth noting that as how multi-modality has secured birds' (and other species) survival in complex environments, similar manifestations in mobile robots can be rewarding and important. 

Payload is a key factor for aerial systems. We are seeking control design paradigms that allow bypassing costly algorithms that depend on heavy on-board computers widely used in legged systems and yet being able to match what they can do using less expensive optimization-free frameworks. In this work, we will present our optimization-free approach based on path planning in the state space of the robot. Our constraints are bounded unilateral ground contact forces which are important to avoid the violation of gait feasibility conditions. The resulting algorithms can run faster and can be hosted by lightweight processors.

From a feedback design standpoint, the challenge of simultaneously providing asymptotic stability and gait feasibility constraints satisfaction in legged systems (bipedal and multi-pedal) has been extensively addressed \cite{apgar_fast_2018,bledt2020regularized,hutter_anymal_2016}. For instance, the works by \cite{westervelt2007feedback} have provided rigorous model-based approaches to assign attributes such as the efficiency of locomotion in off-line fashions. 

Other attempts entail optimization-based, nonlinear approaches to secure safety and performance of legged locomotion \cite{nguyen2015optimal,CLFQP,ames_rapidly_2014,hereid20163d,sentis2006whole}. In general, in these paradigms, an optimization-based controller adjusts the gait parameters throughout the whole gait cycle such that not only the robot's posture is adjusted to accommodate the unplanned posture adjustments but also the joints position, velocity, and acceleration are modified to avoid slipping into infeasible scenarios, e.g., the violation of contact forces. What makes these methods cumbersome is that they are widely defined based on whole-body control at every time-step which can lead to computationally expensive optimization problems. 

Many of today's real-time approaches have achieved better efficiencies through reduced-order models and decomposition methods \cite{buschmann_collocation_2007}, however, they still depend on optimization solvers \cite{koolen_design_2016,fahmi_passive_2019,dario_bellicoso_perception-less_2016,bretl_testing_2008,pardo_evaluating_2016,mastalli_trajectory_2017,mastalli_-line_2015,winkler_gait_2018,aceituno-cabezas_simultaneous_2018,dai_planning_2016}. While many optimization-based methods do not present promising horizons in experimental legged robotics, other Gauss-Newton Hessian approximation such as Sequential Linear Quadratic (SLQ) \cite{sideris_efficient_nodate}, iLQR and iLQG \cite{todorov_generalized_2005} have been more effective. That said, they also present formidable challenges for smaller processors.

Other popular paradigms such as Approximate Dynamic Programming (ADP) \cite{powell2007approximate}, Reinforcement Learning (RL) \cite{sutton2018reinforcement}, decoupled approaches to design control for nonlinear stochastic systems \cite{rafieisakhaei2017near} can potentially remedy the challenges associated with costs of computation. However, these approaches are far from providing any practical solutions to the problem at hand and they are shown to be only effective on simpler practical robots \cite{berseth_progressive_2018,heess_emergence_2017,peng2015learning,xie_feedback_2018,haarnoja_composable_2018,mahmood_benchmarking_2018,haarnoja_learning_2019,haarnoja_reinforcement_2017} and demonstrations in more complex problems such as those offered by legged robots are missing.

In this work, we will resolve gait parameters for our quadrupedal platform off-line and re-plan them in real-time during the gait cycle. We assume well-tuned supervisory controllers found in \cite{sontag1983lyapunov, 371031, bhat1998continuous} and will focus on fine-tuning the joints' desired trajectories to satisfy unilateral contact force constraints. To do this, we will devise intermediary filters based on Explicit Reference Governors (ERG) \cite{411031,bemporad1998reference,gilbert2002nonlinear,dangol2020towards,dangol2020performance,liang2021rough}. ERGs relied on provable Lyapunov stability properties can perform the motion planning problem in the state space in a much faster way than widely used optimization-based methods. 

\section{Overview of Northeastern University's \textit{Husky Carbon} Morpho-functional Platform}

\textit{Husky Carbon} \cite{ramezani2021generative}, shown in Fig.~\ref{fig:husky_overview}, when standing as a quadrupedal robot, is 2.5 ft (0.8 m) tall, is 12 in (0.3 m) wide. The robot was fabricated from reinforced thermoplastic materials through additive manufacturing with a total weight of 9.5 lb (4.3 kg). It hosts on-board power electronics and it operates using an external power supply. The current prototype lacks exteroceptive sensors such as camera and LiDAR. The robot is constructed of two pairs of identical legs in the form of parallelogram mechanisms. Each with three degrees-of-freedom (DOFs), the legs are fixated to Husky's torso by a one-DOF revolute joint with a large range of motion. As a result, the legs can be located sideways. This configuration allows facing the knee actuators upwards for propulsion purposes. A clutch mechanism will disengage the knee actuator from the lower limb before the actuator runs a propeller. 


\section{System dynamic modeling and controller}
\label{sec:modeling}

Husky can be modeled using the simple reduced-order model where the leg is assumed to be massless and all masses are incorporated into the body. Then the system has 6 dynamical DOFs representing the body's linear position and orientation. Each leg is modeled using two hip angles (frontal and sagittal) and a prismatic joint to describe the leg end position. This results in a very simplified model which we will use in a numerical simulation.

\subsection{Husky Reduced Order Model (HROM)}


The Husky Reduced Order Model (HROM) can be derived using Euler-Lagrangian dynamics formulation where the full conformation of Husky as shown in Fig.~\ref{fig:husky_overview} are simplified by assuming massless legs and modeling them as variable-length linkages from the center of mass to the foot end positions. These virtual legs can be defined using two rotations and length variables, and the ground reaction forces (GRF) are applied at the foot end positions which will perturb the system.

Let the superscript $B$ represent a vector defined in the body frame (e.g., $\bm x^B$), and the rotation matrix $R_B$ represents the rotation of a vector from the body frame to the inertial frame (e.g., $\bm x = R_B \bm x^B$). The foot end positions can be derived using the following kinematics equations:
\begin{equation}
\begin{aligned}
    \bm{p}_{F i} &= \bm{p}_{B} + R_{B} \bm{l}_{h i}^{B} + R_{B} \bm{l}_{f i}^{B} \\
    \bm{l}_{f i}^{B} &= R_{y}\left(\phi_{i}\right) R_{x}\left(\gamma_{i}\right)
    \begin{bmatrix} 
    0, & 0, & -r_{i}
\end{bmatrix}^\top,
\label{eq:foot_pos}
\end{aligned}
\end{equation}
where $\bm p_B$ is the body position, $\bm p_{F i}$ is the position of foot $i \in \mathcal{F}$ where $\mathcal{F}$ is the set of legs, $\phi_i$ and $\gamma_i$ are the hip frontal and sagittal angles respectively, and $r_i$ is the prismatic joint length.

Let $\bm \omega_B$ be the body angular velocity in the body frame and $\bm g$ be the gravitational acceleration vector. The legs of HROM are massless, so we can ignore all leg states and directly calculate the kinetic energy $K= (m_{B} \dot{\bm{p}}_{B}^{\top} \dot{\bm{p}}_{B} + \bm{\omega}_{B}^{\top} \hat{I}_{B} \bm{\omega}_{B})/2$ and potential energy $V=-m_{B} \bm{p}_{B}^{\top} \bm{g}$. Then the Lagrangian of the system can be calculated as $L = K - V$ and the dynamic equation of motion can be derived using the Euler-Lagrangian Method. The body orientation is defined using the Hamiltonian's principles and the modified Lagrangian for rotation in SO(3) to avoid using Euler rotations which can become singular during the simulation. Then the equation of motion can be derived as follows:
\begin{equation}
\begin{gathered}
    \textstyle \frac{d}{d t} \left( \frac{ \partial L}{\partial \dot{\bm p}_B } \right ) - \frac{\partial L}{\partial \bm{p}_{B}} = \bm{u}_{1}, \qquad
    \dot{R}_B = R_B\,[\bm \omega_B]_\times \\
    \textstyle \frac{d}{dt}\left( \frac{\partial L}{\partial \bm \omega_B^B}  \right) + 
    \bm \omega_B \times \frac{\partial L}{\partial \bm \omega_B} + 
    \sum_{j=1}^{3} \bm r_{Bj} \times \frac{\partial L}{\partial \bm r_{Bj}} = \bm u_2,
\end{gathered}
\label{eq:euler-lagrangian}
\end{equation}
where $\bm u_1$ and $\bm u_2$ are the generalized vectors, $[\,\cdot\,]_\times$ is the skew operator, and $R_B^\top = [\bm r_{B1}, \bm r_{B2}, \bm r_{B3}]$. The dynamic system acceleration can then be solved from into the following standard form:
\begin{equation}
M\,
\begin{bmatrix}
\ddot{\bm p}_B \\ \dot{\bm \omega}_B
\end{bmatrix} 
+ \bm h = \textstyle \sum_{i=1}^{4} B_{gi}\,\bm u_{gi},
\label{eq:eom_dynamics}
\end{equation}
where $M$ is the mass/inertia matrix, $\bm h$ contains the coriolis and gravitational vectors, $B_{gi} \bm u_{gi}$ represent the generalized force due to the GRF ($\bm u_{gi}$) acting on the foot $i$, and $B_{gi} = (\partial \dot{\bm p}_{Fi}/ \partial \bm v)$ where $\bm v = [\dot{\bm p}_B^\top, \bm \omega_B^\top]^\top$. $M$, $\bm h$, and $B_{gi}$ are also a function of the leg joint variables ($\phi_i$, $\gamma_i$, and $r_i$), which is driven by setting their accelerations to track a desired joint states. The joint states and the inputs are defined as follows
\begin{equation}
    \bm q_L = [\bm \phi^\top, \bm \gamma^\top, \bm r^\top]^\top, \qquad
    \ddot{\bm q}_L = \bm u_L,
\end{equation}
where $\bm \phi$, $\bm \gamma$, and $\bm r$ contain the joint variables of all legs (hip frontal, hip sagittal, and leg length respectively), and $\bm u_L$ forms the control input to the system in the form of the leg joint state accelerations. Combining both the dynamic and massless leg states form the full system states:
\begin{equation}
    \bm x = [\bm q_L^\top, \dot{\bm q}_L^\top, \bm r_B^\top, \bm p_B^\top, \dot{\bm p}_B^\top, \bm \omega_B^\top]^\top,
\end{equation}
where the $\bm r_B$ contains the elements of the rotation matrix $R_B$. The equation of motion can then simply defined as $\dot{\bm x} = \bm f(\bm x, \bm u_L, \bm u_g)$ which will be implemented in a numerical simulation.

The GRF is modeled using a compliant ground model and Stribeck friction model, defined as follows:
\begin{equation}
\begin{aligned}
    \bm u_{gi} & = \begin{cases} \, 0 &  \mbox{if } z_i > 0  \\
    \, [u_{gi,x},\, u_{gi,y},\, u_{gi,z}]^\top & \mbox{else} \end{cases} \\
    u_{gi,z} &= -k_{gz} z_i - k_{dz} \dot{z}_i \\
    u_{gi,x} &= - s_{i,x} u_{gi,z} \, \mathrm{sgn}(\dot{x}_i) - \mu_v \dot{x}_i \\
    s_{i,x} &= \left(\mu_c - (\mu_c - \mu_s) \mathrm{exp} \left(-|\dot{x}_i|^2/v_s^2  \right) \right),
\end{aligned}
\end{equation}
where $x_i$ and $z_i$ are the $x$ and $z$ positions of foot $i$  respectively, $k_{gz}$ and $k_{dz}$ are the spring and damping coefficients of the compliant surface model respectively, $u_{gi,x}$ and $u_{gi,y}$ are the ground friction forces respectively, $\mu_c$, $\mu_s$, and $\mu_v$ are the Coulomb, static, and viscous friction coefficients respectively, and $v_s > 0$ is the Stribeck velocity. We omit the derivations of $u_{g,y}$ as it follows similar derivations to $u_{g,x}$.

\subsection{Controller and Explicit Reference Governor (ERG)}

\begin{figure}[t]
\centering
\vspace{0.05in}
\includegraphics[width=0.7\linewidth]{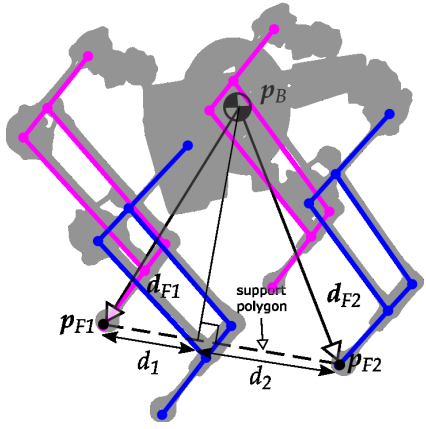}
\caption{The reduced order-model represented by a triangular pendulum which is used to estimate GRF and constraints equations for ERG.}
\vspace{-0.05in}
\label{fig:HROM_tripen}
\end{figure}

The ERG framework is used to enforce the ground force constraint during the implementation of the walking gait. The ERG is utilized by defining a cost or constraint equation as a function of the controller reference. However, calculating the GRF is very complex and difficult even in HROM, so a further simplified model is used to estimate the GRF and to derive the constraint equations.

The ROM is modeled as a triangular inverted pendulum (TIP), shown in Fig.~\ref{fig:HROM_tripen}, which is an extension of the more standard variable length inverted pendulum model. The two contact points at the feet allow us to estimate the distribution of GRF between the two legs so we can evaluate the friction constraint on each leg individually. The kinematic constraints of the TIP model can be defined as follows:
\begin{equation}
\ddot{\bm p}_{F1}=\ddot{\bm{d}}_{F1}+\ddot{\bm{p}}_{B} = 0, \qquad
\ddot{\bm p}_{F2}=\ddot{\bm{d}}_{F2}+\ddot{\bm{p}}_{B} = 0,
\end{equation}
where $\bm p_{F1}$ and $\bm p_{F2}$ are the front and rear foot position respectively and $\bm d_{Fi}$ represent the leg position from the body CoM to either foot. Assuming no slippage at the foot end points ($\ddot{\bm{p}}_{F1} = \ddot{\bm{p}}_{F2} = 0$), the body can be controlled using the leg length acceleration $\ddot{\bm{d}}_{F1}=\ddot{\bm{d}}_{F2} = -\ddot{\bm{p}}_B$. In this case, the body acceleration $\ddot{\bm{p}}_B$ is the control input to this simplified model which will be mapped back into the joint space accelerations ($\bm u_L$) for HROM. Consider the following PD controller to track a desired body states ($\bm p_{B,ref}$) and velocity:
\begin{equation}
\ddot{\bm{p}}_{B} = K_{p} \left( \bm{p}_{B,ref} - \bm{p}_{B} \right) + K_{d} \left(\dot{\bm p}_{B,ref} - \dot{\bm p}_{B} \right),
\label{eq:controller_erg}
\end{equation}
where $K_p$ and $K_d$ are the PD gains for this controller. Define the vectors $\bm x_{p} = [\bm p_B^\top, \dot{\bm p}_B^\top]^\top$ and $\bm x_r = [\bm p_{B,ref}^\top, \dot{\bm p}_{B,ref}^\top]^\top$ which turns \eqref{eq:controller_erg} into the concise form of $\ddot{\bm p}_B = K(\bm x_r - \bm x_p)$ where $K = [K_P, K_d]$.

The ground forces can be derived from the body acceleration, as this point mass dynamics can be modeled as follows:
\begin{equation}
    m_B\,\ddot{\bm p}_B = m_B\,\bm g + \bm u_{g1} + \bm u_{g2},
\label{eq:eom_tip}
\end{equation}
where $\bm u_{g1}$ and $\bm u_{g2}$ are the GRF of the front and rear foot respectively. Since we can only solve for the sum of the ground forces from \eqref{eq:eom_tip}, we need to make some assumptions to solve for $\bm u_{g1}$ and $\bm u_{g2}$. This model restricts moment about the axis perpendicular to the support polygon and by assuming the lateral ground forces are distributed evenly, we can form the following constraint equations:
\begin{equation}
\begin{gathered}
    u_{g1,y} = u_{g2,y}, \qquad
    |d_1|\, u_{g1,z} = |d_1|  \, u_{g2,z} \\
    |d_1|\, \hat{\bm e}_g^\top \, \bm u_{g1} = |d_2|\, \hat {\bm e}_g^\top\, \bm u_{g2},
\end{gathered}
\label{eq:eom_tip_constraints}
\end{equation}
where $\hat{\bm e}_g$ is the unit vector perpendicular to the support polygon along the ground plane, and $d_1$ and $d_2$ are the distance along the support polygon as illustrated in Fig.~\ref{fig:HROM_tripen}. Combining \eqref{eq:eom_tip} and \eqref{eq:eom_tip_constraints} yields the general form equation $A [\bm u_{g1}^\top, \bm u_{g2}^\top]^\top = \bm b$ where the GRF can be solved given that $A$ is invertible. The GRF can be defined as a function of the body position reference by plugging in \eqref{eq:controller_erg} into \eqref{eq:eom_tip} which will be used to derive the constraint equations for the ERG.

\section{Explicit Reference Governor}
\label{sec:erg}

Explicit reference governor (ERG) works as an add-on scheme on a pre-stabilized closed-loop system \cite{garone2015explicit}. The ERG manipulates the applied reference to the controller to enforce the desired constraints while being as close as possible to the desired reference trajectory.

\subsection{Constraint Equation Formulations}

Some constraint equations are defined to enforce the no-slip condition on both stance feet of the robot. Let the constrain equation be defined as follows:
\begin{equation}
    \bm h_r(\bm x_p, \bm x_r) = J_r(\bm x_p) \,\bm x_r + \bm d_r(\bm x_p) \geq 0,
\label{eq:erg_constraints}
\end{equation}
where the relationship between the constraint and $\bm x_r$ is linear. The GRF for both feet can be solved using \eqref{eq:controller_erg} to \eqref{eq:eom_tip_constraints}, as follows:
\begin{equation}
\begin{gathered}
    \begin{bmatrix}
    \bm u_{g1} \\ \bm u_{g2}
    \end{bmatrix} = A^{-1} \begin{bmatrix}
    m_B\, K\,(\bm x_p - \bm x_r) - m_B\, \bm g \\
    \bm 0_{3 \times 1}
    \end{bmatrix},
\end{gathered}
\label{eq:erg_ground_forces}
\end{equation}
which is linear in $\bm x_r$. The following GRF constraints are used to prevent slipping:
\begin{equation}
\begin{gathered}
    |u_{gi,x}| \leq \mu_s\, u_{gi,z}, \qquad
    |u_{gi,y}| \leq \mu_s\, u_{gi,z}, \\
    u_{gi,z} \geq u_{gi,z}^{min}.
\end{gathered}
\label{eq:erg_ground_forces_constraint}
\end{equation}
Then the constraint equations can be formulated as follows:
\begin{equation}
\begin{gathered}
    \begin{bmatrix}
    -\mathrm{sgn}(u_{gi,x}) & 0 & \mu_s \\
    0 & -\mathrm{sgn}(u_{gi,y}) & \mu_s \\
    0 & 0 & 1
    \end{bmatrix} 
    \bm u_{gi} +
    \begin{bmatrix}
    0 \\ 0 \\ -u_{gi,z}^{min}
    \end{bmatrix} \geq 0,
\end{gathered}
\label{eq:erg_no_slip_constraints}
\end{equation}
for leg $i \in {1,2}$. The $J_r$ and $\bm d_r$ in \eqref{eq:erg_constraints} can be calculated numerically using \eqref{eq:erg_ground_forces} and \eqref{eq:erg_no_slip_constraints} which will be used to formulate the ERG applied reference update law.

\subsection{ERG Formulations}

\begin{algorithm}[ht]
\caption{ERG algorithm to constraint GRFs}\label{alg:erg}
\begin{algorithmic}[1]
\State $\bm{h}_{r}=\bm{h}_{r}\left(\bm{x}_p, \bm{x}_{r}\right)$
\State $\bm{h}_{w}=\bm{h}_{r}\left(\bm{x}_p, \bm{x}_{w}\right)$
\State $\bm{v}_{r}=\bm{v}_{t}=\bm{v}_{n}=\mathbf{0}$
\State $C_{r}=[ \quad ]$
\If{$\min \left(\bm{h}_{w}\right) \geq 0 \text { or } \min \left(\bm{h}_{r}\right) \geq 0$}
\State $v_{r}=\alpha_{r}\left(x_{r}-x_{w}\right)$
\EndIf
\If{$\min \left(\bm{h}_{w}\right) \geq 0 \text { and } \min \left(\bm{h}_{r}\right)<0$}
    \State $n_{c}=\operatorname{length}\left(\bm{h}_{r}\right)$
    \For{$k=1: n_{c}$}
        \State $v_{r}=\alpha_{r}\left(x_{r}-x_{w}\right)$
        \If{$h_{r, i}<0$}
            \State $C_{r}=\left[\bm{C}_{r} ; \bm{J}_{r}(k,:)\right]$
        \EndIf
    \EndFor
    \State $\bm{N}_{r}=\operatorname{null}\left(\bm{C}_{r}\right)$
    \State $[\sim, n]=\operatorname{size}\left(\bm{N}_{r}\right)$
    \State $\bm{v}_{t}=\mathbf{0}$
    \For{$k=1: n$}
        \State $\bm{n}_{k}=\bm{N}_{r}(:, k)/ \bm{N}_{r}(:, k)$
        \State $\bm{v}_{t}=\bm{v}_{t}+\alpha_{t} \bm{n}_{k}\bm{n}_{k}^{\top}\left(\bm{x}_{r}-\bm{x}_{w}\right) $
    \EndFor
\EndIf

\If{$\min \left(\bm{h}_{w}\right) < 0 \text { and } \min \left(\bm{h}_{r}\right)<0$}
    \State $k_{min}=\mathop{\arg\min}\limits_{k} \left(\bm{h}_{w}\right)$ \Comment{index of the smallest $h_w$}
    \State $\bm{r}_{k}=\bm{J}_{r}(k_{min},:)/ \bm{J}_{r}(k_{min},:)$
    \If{${h}_{r}(k_{min}) \geq {h}_{w}(k_{min})$}
        \State $\bm{v}_{n}=\alpha_{n} \bm{r}_{k}\bm{r}_{k}^{\top}\left(\bm{x}_{r}-\bm{x}_{w}\right) $
    \Else
        \State $\bm{v}_{n}=-\alpha_{n} \bm{r}_{k}\bm{r}_{k}^{\top}\left(\bm{x}_{r}-\bm{x}_{w}\right) $
    \EndIf
\EndIf

\State $\dot{\bm{x}}_{w}=\bm{v}_{r}+\bm{v}_{t}+\bm{v}_{n}$
\State $\bm{x}_{w}=\bm{x}_{w}+\Delta t \dot{\bm{x}}_{w}$
\end{algorithmic}
\label{alg:ERG}
\end{algorithm}

\begin{figure}[t]
    \vspace{0.05in}
    \centering
    \includegraphics[width=0.7\linewidth]{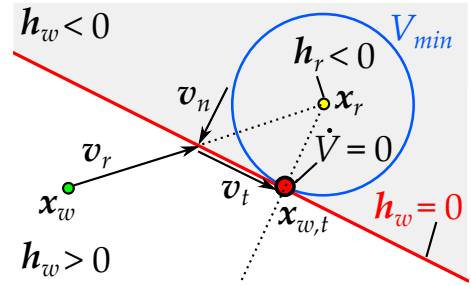}
    \caption{Explicit Reference Governor (ERG) manipulates the applied reference states $\bm x_w$ to be as close as possible to the desired reference $\bm x_r$ without violating the constraint equation $\bm h_w \geq 0$.}
    \label{fig:erg}
    \vspace{-0.05in}
\end{figure}

The ERG framework is utilized to enforce the friction pyramid constraint in \eqref{eq:erg_no_slip_constraints} which is formulated into the form shown in \eqref{eq:erg_constraints}. The ERG algorithm manipulates the applied reference into the controller in \eqref{eq:controller_erg} such that \eqref{eq:erg_constraints} is satisfied. This method is very useful as it avoids using optimization or nonlinear MPC frameworks to enforce constraints on the harder-to-model GRFs.

Let $\bm x_w$ be the applied reference which will be used in the controller instead of $\bm x_r$, and $\bm h_w = \bm h_r(\bm x_p, \bm x_w)$ be the constraint equation value using $\bm x_w$. The ERG manipulates the applied reference ($\bm x_w$) to avoid violating the constraint equation $\bm h_w \geq 0$ while also be as close as possible to the desired reference ($\bm x_r$), as illustrated in Fig.~\ref{fig:erg}. Consider the Lyapunov equation $V = (\bm x_r - \bm x_w)^\top P (\bm x_r - \bm x_w)$. $\bm x_w$ is updated through the update law:
\begin{equation}
    \dot{\bm x}_w = \bm v_r + \bm v_t + \bm v_n,
\label{eq:erg_update}
\end{equation}
where $\bm v_r$ drives $\bm x_w$ directly to $\bm x_r$, while $\bm v_t$ and $\bm v_n$ drives $\bm x_w$ along the surface and into the boundary $\bm h_w =  0$, respectively. The objective of this ERG algorithm is to drive $\bm x_w$ to the state $\bm x_{w,t}$ which is the minimum energy solution $V_{min}$ that satisfies the constraint $\bm h_w \geq 0$. The overall structure of this algorithm can be seen in Algorithm \ref{alg:ERG}.


Let $C_r$ be the rowspace of the violated constraints of $\bm{h}_r$ (rows of the $J_r$ where the constraint is violated). Define $N_r = \mathrm{null}(C_r) = [\bm{n}_1, \dots, \bm{n}_{n}]$ where $n$ is the size of the nullspace. Additionally, let $\bm r_k$ be the $k$'th row of $J_r$. Then the following update law is used for $\bm v_r$, $\bm v_t$, and $\bm v_n$:
\begin{equation}
\begin{gathered}
    \bm v_r = \hat \alpha_r\, (\bm x_r - \bm x_w), \qquad
    \bm v_n = \hat \alpha_n\,\bm r_k\,\bm r_k^\top\, (\bm x_r - \bm x_w) \\
    \bm v_t = \textstyle \sum^n_{k=1} \hat{\alpha}_t\, \bm{n}_k\,\bm{n}_k^\top (\bm{x}_r - \bm{x}_w),
\end{gathered}
\label{eq:erg_update_v}
\end{equation}
where $\hat \alpha$ are scalars defined as follows:
\begin{equation}
\begin{aligned}
    \hat \alpha_r &= 
    \begin{cases}
     \alpha_r, & \text{if } \min(\bm h_w) \geq 0 \text{ or } \min(\bm h_r) \geq 0 \\
     0,      & \text{else} \\
    \end{cases} \\
    \hat \alpha_t &= 
    \begin{cases}
     \alpha_t, & \text{if } \min(\bm h_w) \geq 0 \text{ or } \min(\bm h_r) < 0 \\
     0,      & \text{else} \\
    \end{cases} \\
    \hat \alpha_n &= 
    \begin{cases}
     \alpha_n, & \text{if } \min(\bm h_w) \leq \min(\bm h_r) < 0 \\
     -\alpha_n, & \text{if } \min(\bm h_r) < \min(\bm h_w) < 0 \\
     0,      & \text{else,} \\
    \end{cases}
\end{aligned}
\label{eq:erg_update_alpha}
\end{equation}
where $\alpha$ is a positive scalar which determines the rate of convergence. 

Assuming $\dot{\bm x}_r = 0$ and using the update law defined from \eqref{eq:erg_update_v} and \eqref{eq:erg_update_alpha}, we will have $\dot{V} = - 2(\bm x_r - \bm x_w)^\top\, Q\, (\bm x_r - \bm x_w)$, with $Q = P ( \hat \alpha_r\, I + \textstyle \sum^n_{k=1} \hat \alpha_t\, \bm{n}_k \,\bm{n}_k^\top + \hat \alpha_n\,\bm r_k\,\bm r_k^\top)$. We have the gradient $\dot{V} = 0$ if $\min(\bm h_w) \leq 0$ and $\bm n_k \bot (\bm x_r - \bm x_w)$, while $\dot{V} < 0$ when $\min(\bm h_r) \geq 0$ or when $\min(\bm h_w) \geq 0$. In case both applied reference and target constraints equation are violated, we have $\dot V > 0$ which drives the $\bm x_w$ towards the constraint boundary. This allows the $\bm x_w$ to converge to $\bm x_{w,t}$ which is the minimum energy solution that satisfies $\bm{h}_w \geq 0$ as illustrated in Fig.~\ref{fig:erg}.

\section{Numerical Simulation}
\label{sec:simulation}

\begin{figure}[t]
    \vspace{0.05in}
    \centering
    \includegraphics[width=\linewidth]{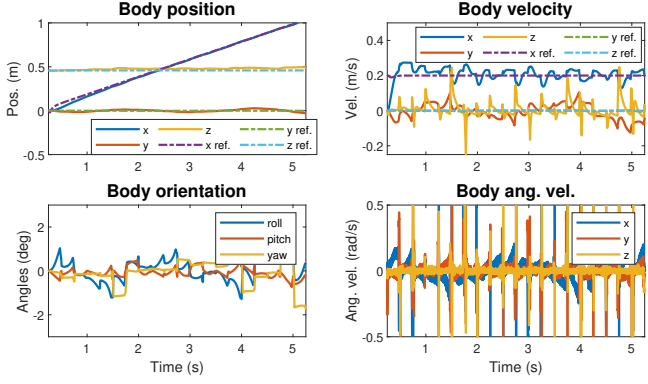}
    \caption{Simulated states of the Husky as it walks stably on a slippery surface ($\mu_s = 0.2$) for 20 walking steps.}
    \label{fig:plot_states}
    \vspace{-0.05in}
\end{figure}

\begin{figure}[t]
    \vspace{0.05in}
    \centering
    \includegraphics[width=\linewidth]{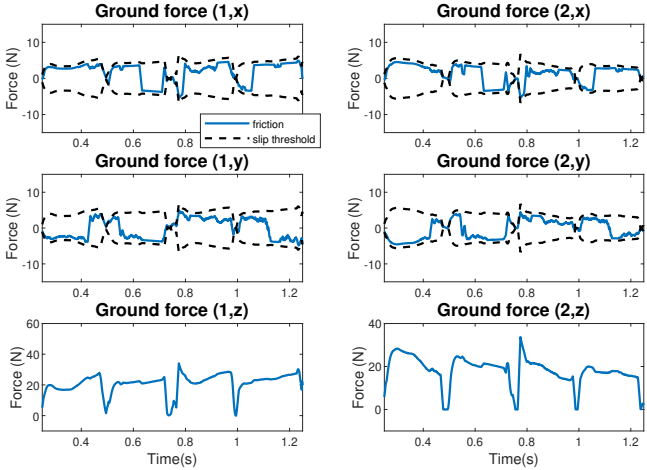}
    \caption{The simulated GRF on the front and rear stance foot, labeled as 1 and 2 respectively, during the first four gait periods. The ground friction stayed within the friction pyramid constraints which are shown as the dashed lines.}
    \label{fig:plot_grf}
    \vspace{-0.05in}
\end{figure}

\begin{figure}[t]
    \vspace{0.05in}
    \centering
    \includegraphics[width=\linewidth]{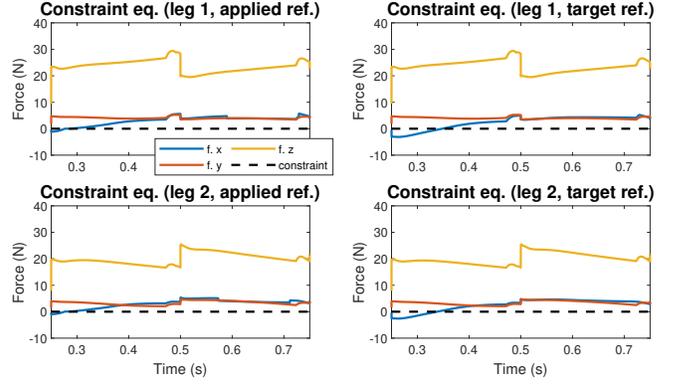}
    \caption{The constraint equation values between using the applied vs. target reference during the first two gait periods. The initial reference cause constraint violation which is avoided using the ERG update law.}
    \label{fig:plot_constraints}
    \vspace{-0.05in}
\end{figure}

\begin{figure}[t]
    \vspace{0.05in}
    \centering
    \includegraphics[width=\linewidth]{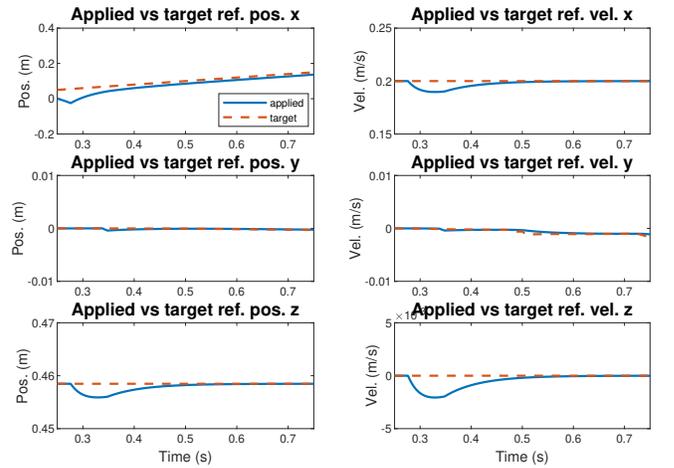}
    \caption{The target vs applied references (body linear position and velocity) during the period of constraint violation of the target references. The applied reference is manipulated to avoid breaking the constraints.}
    \label{fig:plot_references}
    \vspace{-0.05in}
\end{figure}


The simulation was implemented on HROM based on a predefined quasi-static trotting gait in MATLAB. The stance-swing pair switching was done by using a timing-based state machine, where the gait period is defined as 0.25 s and a total of 20 walking steps was simulated. The input of the ERG and the stance foot controller in \eqref{eq:controller_erg} is defined as the inertial frame foot end acceleration relative to the current body position. To implement this controller, we simply integrated this acceleration to the desired position and velocity about the body frame, and track these positions using a PID controller. The swing foot trajectory follows a simple bezier curve trajectory with an end position located at a fixed position relative to the body (0.08 m ahead of hip position in the x-axis). The bezier curve trajectory initial position is set as the previous stance foot final position and a target swing height of 0.2 m with zero initial and final velocity. Since we have no row, pitch, or yaw compensation, we rotate the reference by $R_B^{\top}$ at every time step so that the robot acts like walking on a flat surface without angular deviation.

The simulation results of the robot walking on slippery surface ($\mu_s = 0.2$) at the target forward speed of 0.2 m/s is shown in Fig.~\ref{fig:plot_states} to \ref{fig:plot_references}. The resulting gait is stable, with a potential slipping happening at the beginning of the gait which is regulated by the ERG as shown in Fig.~\ref{fig:plot_grf} and \ref{fig:plot_references}. The ERG has successfully manipulated the controller reference states to avoid constraint violation which resulted in a stable trotting gait even on a slippery surface. Figure~\ref{fig:plot_grf} shows the actual GRF applied to this robot during this simulation and the friction is kept within the constraint boundary defined in \eqref{eq:erg_ground_forces_constraint}.

This simulation was performed in Matlab using a computer with AMD Ryzen six-core 3.40 GHz processor and 16 GB of RAM. Each simulation time step which includes the ERG computation and ODE integration with 4th order Runge-Kutta algorithm was computed at a rate of approximately 2 kHz. Most embedded systems are programmed in C or C++ which are significantly faster than Matlab. This shows that our ERG algorithm can be utilized in real-time computations using the robot's on-board computer.






\section{Conclusions and future work}
\label{sec:conclusion}

We explored the performance of the ERG algorithm on a quadruped where the no-slip constraints have been enforced successfully. The use of ERG allows the robot to walk stably on a slippery surface ($\mu_s = 0.2$) where the controller states are manipulated to satisfy the GRF constraints we specified. This algorithm can be computed quickly, which means that it can serve as an alternative to the more computationally expensive algorithms for enforcing constraints, such as optimization and whole-body control.

The controller considered in this work is only defined for the stance foot where the GRF is applied. Future work should include the usage of the swing foot placement algorithm to regulate the robot heading and the tracking of more complex trajectories. Additionally, other constraints can also be considered for the ERG, such as input and state saturation constraints which are practical for the implementation in the actual robot.






\IEEEtriggeratref{38} 
\bibliographystyle{IEEEtran}
\bibliography{references.bib}

\end{document}